\documentclass[11pt]{article}

\usepackage[preprint]{acl}

\usepackage{times}
\usepackage{latexsym}

\usepackage[T1]{fontenc}

\usepackage[utf8]{inputenc}

\usepackage{devanagari}

\usepackage{microtype}

\usepackage{inconsolata}

\usepackage{graphicx}

\usepackage{float}
\usepackage{placeins}
\usepackage{seqsplit}

\title{A Comparative Analysis of Retrieval-Augmented Generation Techniques for Bengali Standard-to-Dialect Machine Translation Using LLMs}

\author{K. M. Jubair Sami, Dipto Sumit, Ariyan Hossain, Farig Sadeque \\
  Department of Computer Science and Engineering \\
  BRAC University, Dhaka, Bangladesh \\
  \href{mailto:km.jubair.sami@g.bracu.ac.bd}{\{km.jubair.sami}, 
  \href{mailto:dipto.sumit@g.bracu.ac.bd}{dipto.sumit\}@g.bracu.ac.bd} \\
  \href{mailto:ariyan.hossain@bracu.ac.bd}{\{ariyan.hossain}, 
  \href{mailto:farig.sadeque@bracu.ac.bd}{farig.sadeque\}@bracu.ac.bd}
}

\begin{document}
\maketitle
\begin{abstract}
Translating from a standard language to its regional dialects is a significant NLP challenge due to scarce data and linguistic variation, a problem prominent in the Bengali language. This paper proposes and compares two novel RAG pipelines for standard-to-dialectal Bengali translation. The first, a Transcript-Based Pipeline, uses large dialect sentence contexts from audio transcripts. The second, a more effective Standardized Sentence-Pairs Pipeline, utilizes structured local\_dialect:standard\_bengali sentence pairs. We evaluated both pipelines across six Bengali dialects and multiple LLMs using BLEU, ChrF, WER, and BERTScore. Our findings show that the sentence-pair pipeline consistently outperforms the transcript-based one, reducing Word Error Rate (WER) from 76\% to 55\% for the Chittagong dialect. Critically, this RAG approach enables smaller models (e.g., Llama-3.1-8B) to outperform much larger models (e.g., GPT-OSS-120B), demonstrating that a well-designed retrieval strategy can be more crucial than model size. This work contributes an effective, fine-tuning-free solution for low-resource dialect translation, offering a practical blueprint for preserving linguistic diversity.
\end{abstract}

\section{Introduction}

The Bengali language's diverse and culturally significant regional dialects \citep{10796843,khandaker2025bridging, wasi2024diaframe} are critically underrepresented in machine translation (MT) \citep{khandaker2025bridging}. While some research translates dialects into standard Bengali \citep{faria2023vashantor}, the reverse task: translating from standard to regional variants, remains a more challenging and largely unexplored problem \citep{khandaker2025bridging}. This gap is driven by the lack of parallel standard-to-dialect corpora, a common challenge for low-resource languages \citep{klementiev-etal-2012-toward, yakhni-chehab-2025-llms}. Consequently, Large Language Models (LLMs) often fail to capture subtle dialectal nuances without specialized guidance, resulting in inaccurate translations \citep{yakhni-chehab-2025-llms,kadaoui-etal-2023-tarjamat}.

This paper's contributions are as follows:
\begin{itemize}
    \item We design and evaluate two distinct, fine-tuning-free pipelines for standard-to-dialectal Bengali translation using in-context learning: (1) a Transcript-Based Pipeline that uses dialectal audio transcripts as context for large LLMs, and (2) a Standardized Sentence-Pairs Pipeline that uses standard-dialect sentence pairs for smaller LLMs.

    \item We systematically compare these approaches across multiple Large Language Models (LLMs) and six underrepresented Bengali dialects: Chittagong, Comilla, Habiganj, Rangpur, Sylhet, and Tangail.

    \item Our findings identify the optimal strategy for different conditions, providing a practical blueprint for developing Machine Translation (MT) systems for low-resource dialects.
\end{itemize}

\section{Related Works}

The task of translating between standard languages and dialects presents significant challenges. To address these, our work is situated at the intersection of existing research on dialect processing and emerging methodological approaches, which we review below.

\textbf{Bangla Dialect Processing.} Research in Bangla dialect processing has largely focused on identification and dialect-to-standard translation, leveraging the Vashantor dataset \citep{faria2023vashantor}. This corpus, covering the Chittagong, Noakhali, Sylhet, Barishal, and Mymensingh dialects, has been used to train fine-tuned models and prompt LLMs for these tasks \citep{faria2023vashantor, 10796843}. In contrast, the data-scarce standard-to-dialect direction is less explored. This reverse task was addressed by \citet{khandaker2025bridging} via fine-tuning neural models on the Vashantor dataset. Our work also tackles this challenge, proposing a fine-tuning-free, retrieval-augmented alternative.

\textbf{Dialect Translation in Other Languages.} Similar challenges exist elsewhere; studies on Arabic explore fine-tuning and prompting for dialect translation \citep{alabdullah2025advancingdialectalarabicmodern}. Research on Lebanese Arabic highlights LLMs' failure to capture cultural nuances without authentic data \citep{yakhni-chehab-2025-llms}, underscoring our transcript-based approach and the practice of comparing different methods. \citep{alabdullah2025advancingdialectalarabicmodern, liu2023pre, Han2024ClinicalNMT}.

\textbf{Applying RAG to Dialect Translation.} Our pipelines utilize Retrieval-Augmented Generation (RAG), where retrieved dialect sentence pairs or transcript excerpts serve as few-shot in-context examples for an LLM. While RAG is established for question-answering, its application to low-resource dialect translation is an emerging area\citep{perak-etal-2024-incorporating, Ndimbo2025, kyslyi-etal-2025-vuyko, miyagawa-2025-rag}. This RAG-inspired approach mitigates data scarcity and helps preserve culturally specific lexical and pragmatic patterns during translation.

\section{Methodology}
We compare two strategies for standard-to-dialectal Bangla translation, using two distinct datasets each tailored to a specific pipeline.
\subsection{Datasets}
\subsubsection{Dataset-01: Transcript-Based Dataset (for Pipeline 1)}

\begin{table}[H]
  \centering
  \small
  \begin{tabular}{lr}
    \hline
    District & \# of data points \\
    \hline
    Sylhet & 7,624 \\
    Kishoreganj & 2,049 \\
    Narail & 1,859 \\
    Chittagong & 1,757 \\
    Narsingdi & 1,373 \\
    Sandwip & 1,310 \\
    Rangpur & 1,298 \\
    Tangail & 1,271 \\
    Habiganj & 1,170 \\
    Barishal & 1,006 \\
    Comilla & 318 \\
    Noakhali & 278 \\
    \hline
    Total & 21,313 \\
    \hline
  \end{tabular}
  \caption{Dialect coverage and number of sentences in the transcript-based dataset.}
  \label{tab:transcript_dataset}
\end{table}

This dataset \citep{hassan2025regspeech12regionalcorpusbengali}, from transcribed audio of local Bengali dialects, contains long, contextually rich sentences reflecting spoken language. Its broad district coverage captures diverse lexical and syntactic variation, ideal for in-context retrieval by large LLMs.

\subsubsection{Dataset-02: Standardized Sentence-Pairs Dataset (for Pipeline 2)}
Structured as key-value pairs of local\_dialect\_sentence:standard\_bengali\_translation \citep{hassan2025regspeech12regionalcorpusbengali}, the raw data initially contained many small, fragmented sentences. Our preprocessing attempt to merge them yielded modest improvement in similarity search performance.

\begin{table}[H]
  \centering
  \small
  \resizebox{\columnwidth}{!}{%
  \begin{tabular}{lrr}
    \hline
    District & Before preprocessing & After preprocessing \\
    \hline
    Chittagong   & 7,193 & 7,295 \\
    Habiganj     & 5,375 & 5,457 \\
    Rangpur      & 4,061 & 4,140 \\
    Kishoreganj  & 3,653 & 3,898 \\
    Tangail      & 353   & 365   \\
    \hline
    Total        & 20,635 & 21,155 \\
    \hline
  \end{tabular}%
  }
  \caption{Dialect coverage and number of sentence-pairs in the standardized dataset, shown before and after preprocessing.}
  \label{tab:standard_pairs_preproc}
\end{table}

\subsection{Dataset Preprocessing and Indexing}

We developed two distinct preprocessing and indexing pipelines for our retrieval systems to accommodate significant differences between our datasets: the first dataset contains long, formal sentences (mean 38.2 words), while the second has short, conversational fragments (mean 6.9 words), necessitating a more intensive and specialized preprocessing approach.

  

\subsubsection{Pipeline 1: Standard Preprocessing}
For Dataset-01, our pipeline focused on robust cleaning and direct embedding. The key steps were:
\noindent \textbf{Text Cleaning:} We loaded raw transcriptions, filtered invalid data, and ensured consistent UTF-8 encoding.

\noindent \textbf{Metadata and Quality Metrics:} Each entry was augmented with metadata (ID, dialect, length) and quality metrics like word count and text complexity.

\noindent \textbf{Hybrid Indexing:} We adopted a hybrid retrieval approach for both semantic and lexical matching:

\noindent \hspace{1em} \textbf{Dense Index:} We generated 768-dimensional embeddings using the l3cube-pune/bengali-sentence-similarity-sbert model \citep{deode-etal-2023-l3cube}, a model specifically fine-tuned for semantic similarity on Bengali text. These were L2-normalized and indexed with FAISS\footnote{https://github.com/facebookresearch/faiss} (IndexFlatIP) for efficient cosine similarity search \citep{douze2024faiss}.

\noindent \hspace{1em} \textbf{Sparse Index:} Concurrently, we built a rank\_bm25 index for keyword-based sparse retrieval \citep{robertsonbm25}.

\subsubsection{Pipeline 2: Augmented Preprocessing for Short Texts}
The shorter sentence pairs in Dataset-02 required a more sophisticated pipeline to enrich contextual information before embedding.  

\noindent \textbf{Systematic Text Normalization:} We applied a multi-step normalization function including Unicode NFC, standardization of Bengali digits and punctuation, and collapsing repeated whitespace and characters.

\noindent \textbf{Short Fragment Augmentation:} To add crucial context to short texts, we tagged sentences with fewer than three tokens as [[SHORT]] and applied content-based tags like [[QUESTION]]. Consecutive short entries from the same dialect were merged into a single, contextually rich record marked [[MERGED]].

\noindent \textbf{Structured Representation:} Before embedding, each entry was formatted as: District: \{district\} | STANDARD: \{standard\_norm\} | LOCAL: \{local\_norm\_tagged\}. This structure explicitly provides the model with dialectal, standard, and augmented local information to learn region-specific translation patterns.

\noindent \textbf{Hybrid Indexing:} As in the first pipeline, we generated hybrid dense (FAISS) and sparse (BM25) indices from these structured representations to enhance retrieval performance.

The intensive augmentation step was designed to address the fact that shorter sentences in Dataset-02 lack self-contained context. Our merging and tagging strategies artificially created this context, providing a richer signal to the embedding model and mitigating the ambiguity of short utterances.

\subsection{Translation Pipelines}

\subsubsection{Pipeline 1: Transcript-Based Pipeline for Larger LLMs}
This pipeline is designed for simplicity and is particularly effective for large, powerful LLMs that have been pre-trained on extensive Bengali data. The workflow is as follows:

\begin{figure}[t]
  \centering
  
  \includegraphics[width=\columnwidth]{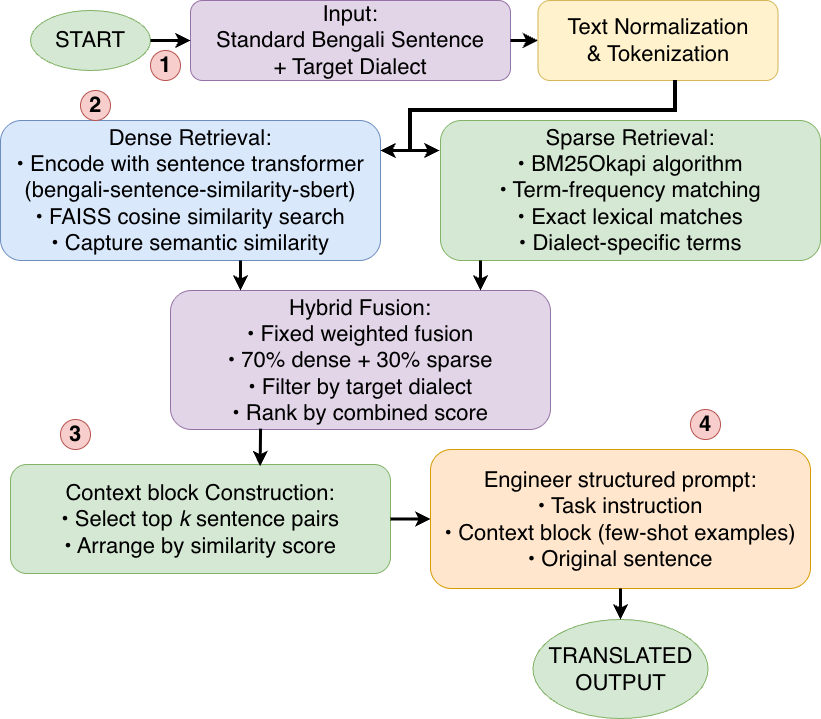}
  \caption{Pipeline 1 translation workflow.}
  \label{fig:pipeline1 flowchart}
\end{figure}

\noindent \textbf{Input:} A standard Bengali sentence and target dialect are provided by the user. The input sentence then undergoes standard text normalization and tokenization to prepare it for processing.

\noindent \textbf{Hybrid Vector-Based Retrieval:} We use a hybrid system to find relevant examples, combining two methods. For \textbf{Dense Retrieval}, the same sentence transformer from indexing generates an embedding of the input to find semantically similar sentences via a cosine similarity search on the FAISS index. For \textbf{Sparse Retrieval}, a BM25Okapi algorithm performs term-frequency based matching to identify sentences with exact lexical matches and key dialect-specific terms. Finally, a \textbf{Hybrid Fusion} combines the scores using a weighted fusion (70\% dense, 30\% sparse), and the results are then filtered by the target dialect and ranked.

\noindent \textbf{Context Construction \& LLM-Based Translation:} A few-shot context is constructed by selecting the top \( n \) (a user-defined hyperparameter) sentence pairs, ranked by similarity. This context, along with a task instruction and the input sentence, is then formatted into a prompt and fed to an LLM to generate the final translation. A sample prompt is provided in Appendix~\ref{sec:sample_prompts}.

\subsubsection{Pipeline 2: Standardized Sentence-Pairs Pipeline for Smaller LLMs}
This pipeline is more complex, designed to maximize retrieval accuracy as Dataset-02 has relatively smaller sentence pairs. Since it retrieves both the local\_dialect and standard\_bengali sentence pairs, it is also designed for smaller, more efficient LLMs that might not be pre-trained on extensive Bengali data. The workflow is as follows:

\begin{figure}[t]
  \centering
  
  \includegraphics[width=\columnwidth]{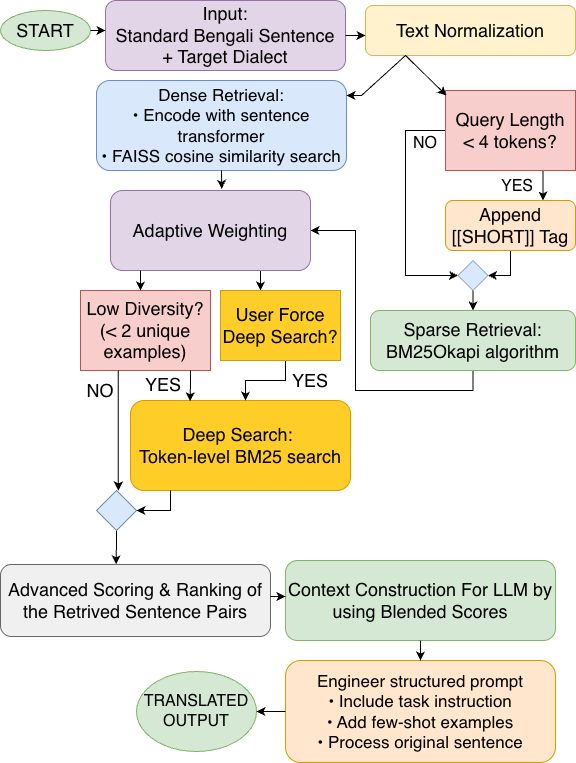}
  \caption{Pipeline 2 translation workflow.}
  \label{fig:sentence_length_histogram}
\end{figure}

\noindent \textbf{Input and Normalization:} The input Standard Bengali sentence undergoes a comprehensive normalization process, which includes Unicode normalization, removal of zero-width characters, punctuation standardization, and numeral conversion. Queries shorter than four tokens are tagged as short.

\noindent \textbf{Hybrid Retrieval with Adaptive Weighting:} We identify relevant sentence pairs using a hybrid approach with dynamic weights. For \textbf{Dense Retrieval}, the same sentence transformer from indexing encodes the input for a FAISS cosine similarity search to find semantically similar examples. \textbf{Sparse Retrieval} uses BM25 for lexical matching. The appended [[SHORT]] tag to the input is to specifically target other short examples in the corpus. The fusion employs \textbf{Adaptive Weighting} based on query length: standard queries favor dense retrieval (55/35), while short queries prioritize sparse retrieval (35/55) to better capture lexical matches. Furthermore, the number of candidates retrieved is doubled for both sparse (50 to 200) and dense (50 to 100) searches to cast a wider net for short queries.

\noindent \textbf{Deep Search for Low-Diversity Queries:} A "Deep Search" mechanism is initiated either automatically when initial results lack diversity (e.g., fewer than two unique examples) or manually by the user. It runs a BM25 search for each input token, aggregates the scores, and re-weights to favor sparse retrieval.

\noindent \textbf{Advanced Scoring and Ranking:} Candidates from the retrieval stages are ranked using a blended score. This final score incorporates the weighted dense and sparse similarity scores, along with several bonuses, including a district matching bonus, significant bonuses for exact and substring matches, and a minor bonus based on character-level similarity.

\noindent \textbf{Context Construction \& LLM-Based Translation:} A few-shot context is constructed by filtering top \( n \) (a user-defined hyperparameter) ranked sentence pairs by the target dialect, and sorting by score. A prompt containing these standard\_bengali:local\_dialect examples, along with instructions and the input sentence, is then sent to an LLM to generate the final translation. A sample prompt is provided in Appendix~\ref{sec:sample_prompts}. 

\section{Experiments and Results}
\subsection{Experimental Setup}
To investigate the relationship between model characteristics and pipeline design in dialectal translation, we evaluated our pipelines across a diverse set of LLMs, ranging from smaller open-weight models to larger ones, as well as proprietary models. The comparison covered six Bengali dialects: Chittagong, Habiganj, Rangpur, Tangail (present in both datasets), and Comilla and Sylhet (only in Dataset-01). We assessed translation quality using complementary metrics covering lexical overlap (BLEU \citep{papineni-etal-2002-bleu}, ChrF \citep{popovic-2015-chrf}), edit distance (WER), and learned semantic similarity (BERTScore F1 \citep{Zhang*2020BERTScore:}), evaluated on $N=50$ diverse sentence pairs per dialect, totaling 7,700 data points across all pipeline-dialect combinations. Detailed metric formulations and implementation specifics are provided in Appendix~\ref{sec:metrics_details}.


\section{Results and Analysis}
\begin{figure}[H]
  \centering
  \includegraphics[width=\columnwidth]{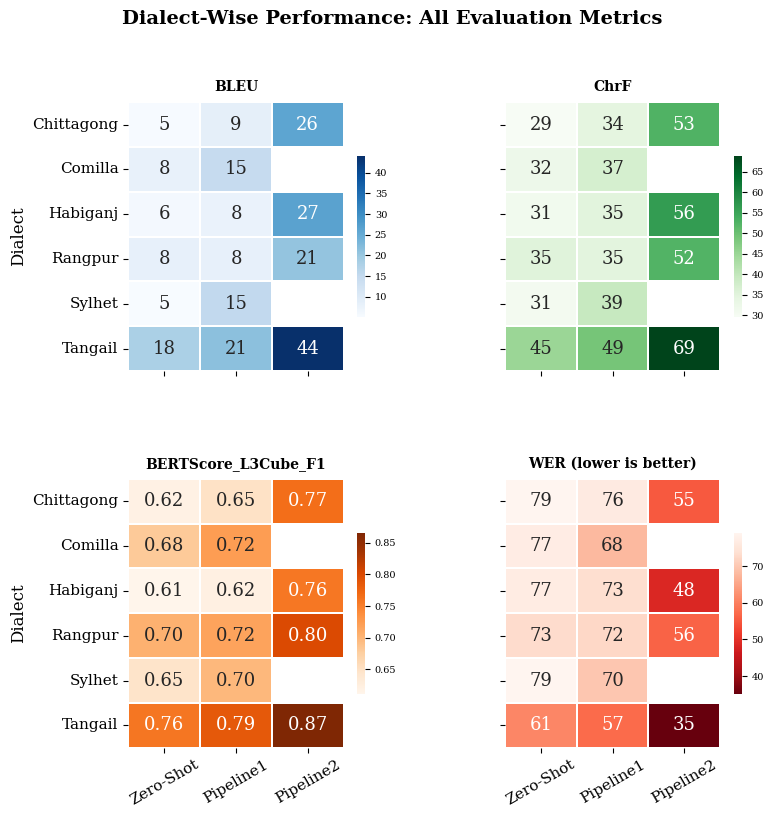}
  \caption{Dialect-wise performance comparison across zero-shot, Pipeline 1, and Pipeline 2 settings, with scores averaged across all LLMs. The hierarchy is clear: Pipeline 2 > Pipeline 1 > Zero-shot.}
  \label{fig:heatmap_all_appendix}
\end{figure}

We evaluated both pipelines across multiple LLMs and six Bengali dialects. Figure~\ref{fig:heatmap_all_appendix} presents a comprehensive performance overview, with scores averaged across all LLMs for each pipeline-dialect combination. It is important to note that this averaging can sometimes mask the peak performance of the best models, as lower-performing models can pull down the aggregate score. Complete performance tables showing individual LLM results for all dialects are provided in Appendix~\ref{sec:complete_tables}. Nevertheless, Pipeline 2 consistently outperforms Pipeline 1, a difference largely attributable to its superior data structure and preprocessing. This performance gap is also qualitatively evident in the prompts themselves. As illustrated in Appendix~\ref{sec:sample_prompts} using a consistent example sentence, the structured few-shot pairs in Pipeline 2 produce highly accurate translations, whereas Pipeline 1 only partially captures dialectal nuances and the Zero-Shot baseline fails entirely. Our quantitative analysis also shows that dialectal proximity to Standard Bengali strongly correlates with translation quality, and well-designed RAG pipelines enable smaller models to compete with larger ones. Detailed model-wise comparisons are shown in Appendix~\ref{sec:detailed_results}. A comparison with \citet{khandaker2025bridging}'s fine-tuned models is provided in Appendix~\ref{sec:khandaker_comparison}.

\subsection{Pipeline Comparison}
As shown in Figure~\ref{fig:heatmap_all_appendix}, Pipeline 2 systematically outperforms Pipeline 1 across all shared dialects (e.g., Chittagong: BLEU 9→26, WER 76\%→55\%). This stems from Dataset-02's explicit local\_dialect:standard\_bengali pairs providing ideal few-shot context versus Dataset-01's raw transcripts, plus significantly higher number of data points (Chittagong: 7,295 vs. 1,757 examples) with advanced preprocessing for short fragments.

\subsection{Linguistic Proximity Dominates}
Dialectal similarity to Standard Bengali is the strongest performance predictor. Tangail achieves the highest scores (BLEU=44, WER=35) with only 365 examples, while divergent dialects like Chittagong (WER=55) and Sylhet/Comilla (WER=70/68) require both abundant data and intensive preprocessing. Intermediate dialects (Habiganj, Rangpur: WER=48/56) show moderate divergence can be mitigated via Pipeline 2. Critically, this linguistic proximity advantage persists even in zero-shot scenarios: Tangail achieves BLEU=18 and WER=61 without any dialectal examples, outperforming divergent dialects Chittagong (BLEU=5, WER=79) and Sylhet (BLEU=5, WER=79) by 3.4-3.6× in BLEU scores, confirming that inherent linguistic similarity to Standard Bengali remains the dominant factor regardless of learning paradigm.

\subsection{LLM Performance}
Zero-shot translation consistently fails (BLEU=5-12, WER=67-84\%). Pipeline 2 enables dramatic gains: Gemma-3-27B improves from WER=76.62\% to 36.70\%, achieving best overall performance (BLEU=45.06). Critically, smaller models like Llama-3.1-8B (WER=51.18) outperform much larger models like GPT-OSS-120B (WER=52.65), demonstrating retrieval quality can compensate for model capacity (Appendix~\ref{sec:model_comparison}).

\section{Conclusion and Future Work}
We proposed two RAG-based pipelines for standard-to-dialectal Bengali translation. Pipeline 2 (Standardized Sentence-Pairs) proves most effective, enabling smaller models to outperform higher-parameter counterparts by converting an intractable zero-shot task into a manageable few-shot problem. While linguistic proximity to Standard Bengali strongly correlates with performance, our fine-tuning-free approach provides a practical blueprint for low-resource dialect translation. This work is ongoing: we are actively expanding Dataset-02, developing a more optimized version of Pipeline 2, and investigating fine-tuning-based approaches alongside retrieval-augmented methods.

\section*{Limitations and Challenges}
Despite the promising results, this study is subject to several limitations and faced inherent challenges that warrant discussion.

\begin{itemize}
    \item \textbf{Data Availability and Quality:} The primary challenge remains the scarcity of high-quality, parallel corpora for Bengali dialects. While our pipelines aimed to mitigate this, their performance is still fundamentally constrained by the volume and cleanliness of the underlying datasets. The datasets used contained inconsistencies and noise inherent to transcribed spoken language, which could impact retrieval accuracy.
    \item \textbf{Limited Dialectal Coverage:} Our evaluation was confined to six Bengali dialects. Given the vast number of dialects spoken across Bangladesh and West Bengal, our findings may not be generalizable to all linguistic variants, especially those with more pronounced structural differences from Standard Bengali.
    \item \textbf{Evaluation Constraints:} Our evaluation was constrained by limited time, computational resources, and the availability of human annotators. Consequently, we utilized a curated test set of $N=50$ diverse sentence pairs per dialect. Across all combinations of pipelines and dialects, this amounted to a total of 7,700 data points. While curated for diversity, this sample size is a limitation; more robust results would require a larger test set. A second major limitation is our exclusive reliance on automated metrics. These metrics fail to capture critical nuances of dialectal appropriateness, fluency, and cultural context, which can only be assessed through human evaluation.
    \item \textbf{Absence of a Production-Ready Baseline:} A direct comparison with fine-tuned models on the exact same standard-to-dialect task was not performed within this study's scope. While \citet{khandaker2025bridging} explored fine-tuning using smaller, neural models, a head-to-head comparison would be needed to precisely quantify the trade-offs between RAG with LLMs and fine-tuned smaller models.
\end{itemize}

\section*{Ethical Considerations}
Developing technology for low-resource dialects carries significant ethical responsibilities. While this work aims to support linguistic diversity, it is crucial to consider the potential impacts.
\begin{itemize}
    \item \textbf{Preservation vs. Misrepresentation:} The goal is to preserve and promote dialectal use. However, inaccurate or culturally insensitive translations generated by automated systems risk misrepresenting the language and its speakers. There is a danger of propagating stereotypes or producing nonsensical text that could undermine the perceived value of the dialect.
    \item \textbf{Data Sovereignty and Consent:} The datasets used in this research were drawn from existing public collections. Future data collection efforts must prioritize ethical practices, including obtaining informed consent from native speakers, ensuring fair compensation for their linguistic expertise, and respecting community ownership of the data.
    \item \textbf{Inadvertent Standardization:} The creation of translation tools, by its nature, involves a degree of standardization. There is a risk that such tools could inadvertently promote a single, computationally convenient version of a dialect, thereby eroding the rich, organic micro-variations that exist within dialect communities. Engagement with linguists and community members is vital to mitigate this risk.
    \item \textbf{Usage of AI Tools:} We acknowledge the use of AI-based writing assistants in the preparation of this paper for improving grammar and style. The core ideas, experimental design, and analysis were conducted entirely by the authors.
\end{itemize}





\bibliography{custom}

\appendix



\section{Evaluation Metrics: Detailed Formulations}
\label{sec:metrics_details}

To provide a comprehensive assessment of translation quality, we report a set of complementary metrics that cover lexical overlap, edit distance, and learned semantic similarity. All metrics were evaluated on $N=50$ sentence pairs per dialect, covering a wide range of Bengali lexical diversity.

\begin{enumerate}
  \item \textbf{Corpus-level overlap metrics (BLEU and ChrF)}: We report BLEU \citep{papineni-etal-2002-bleu} and ChrF \citep{popovic-2015-chrf} as corpus-level overlap metrics. Both metrics are computed at the corpus level by aggregating their underlying counts (for example, n-gram matches and candidate/reference lengths) across all sentences before applying the final scoring formula. Aggregating counts prior to final computation preserves the correct statistical behavior of these metrics and avoids inflation that can occur when averaging sentence-level scores on small test sets. Formally, let $M$ denote either BLEU or ChrF, and let Numerator$_i$ and Denominator$_i$ be the per-sentence internal counts used by $M$. The corpus score is calculated by aggregating those counts across all sentences and then applying the metric's scoring function:
  \begin{equation}
    \scalebox{0.7}{$M_{\mathrm{corpus}} = M\Big(\sum_{i=1}^N \mathrm{Numerator}_i,\;\sum_{i=1}^N \mathrm{Denominator}_i\Big).$}
  \end{equation}
  
  \item \textbf{Edit-distance metric (WER)}: Word Error Rate (WER) measures the minimum number of substitutions (S), deletions (D) and insertions (I) needed to convert a hypothesis into a reference, normalized by the reference length. For a robust corpus-level estimate we weight each sentence's WER by its reference word count (RefWC$_i$) and compute the length-weighted average:
  \begin{equation}
    \mathrm{WER}_{\mathrm{corpus}} = \frac{\sum_{i=1}^N \mathrm{WER}_i \times \mathrm{RefWC}_i}{\sum_{i=1}^N \mathrm{RefWC}_i},
  \end{equation}
  where WER$_i$ is the sentence-level WER for sentence $i$.

  \item \textbf{Learned semantic similarity (BERTScore F1)}: BERTScore F1 \citep{Zhang*2020BERTScore:} computes a soft token-level alignment using contextual embeddings (we used the L3Cube Bengali variant to generate embeddings to calculate BERTScores) and produces a continuous similarity score per sentence. As a learned metric, BERTScore is far more robust than $n$-gram methods (BLEU) at capturing semantic equivalence, which is particularly valuable for evaluating low-resource languages like Bengali where lexical variation is common. Because BERTScore is designed as a sentence-level metric, we report the final corpus-level BERTScore as the arithmetic mean of the $N$ sentence scores:
\begin{equation}
  M_{\mathrm{corpus}} = \frac{1}{N}\sum_{i=1}^N M_i,
\end{equation}
where $M_i$ is the BERTScore F1 of sentence $i$.

  \item \textbf{Implementation details}: All metrics were computed using standard, publicly available implementations with default settings unless otherwise noted. In particular, BLEU and ChrF were computed using corpus-level aggregation (not averaged segment BLEU/ChrF), WER was computed using a standard minimum-edit-distance alignment and length-weighted aggregation, and BERTScore F1 was computed at the sentence level and averaged across the evaluation set.
\end{enumerate}

\section{Illustrative Prompt Examples and Translation Quality}
\label{sec:sample_prompts}

This section presents a comparative analysis of prompts from our three experimental setups: Zero-Shot, Pipeline 1 (Transcript-Based), and Pipeline 2 (Standardized Sentence-Pairs). To clearly demonstrate the impact of each prompting strategy, we use the same standard Bengali input sentence, target dialect, and LLM in all examples. The resulting translations highlight a clear progression in quality: the Zero-Shot approach completely fails to capture dialectal features, Pipeline 1 partially captures them, and Pipeline 2 produces the most accurate and fluent dialectal output.

\subsection{Zero-Shot: Prompt Example}
\label{sec:pipeline1_prompt}

\begin{figure}[H]
  \centering
  \includegraphics[width=\columnwidth]{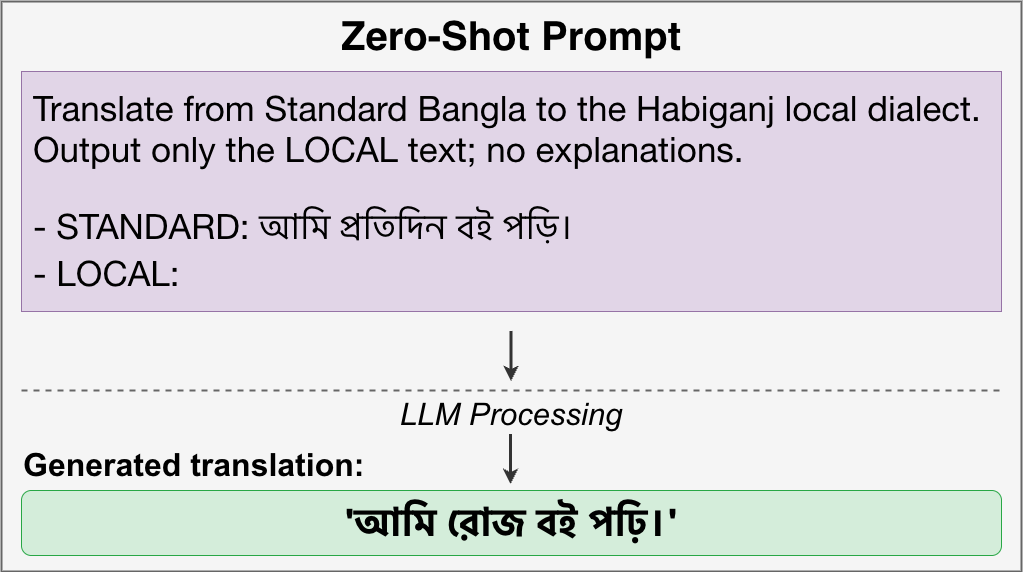}
  
  \caption{A sample prompt that used while translating a sentence in Zero-Shot scenarios.}
  \label{fig:zero_shot_prompt}
\end{figure}

\subsection{Pipeline 1: Transcript-Based Prompt Example}
\label{sec:pipeline1_prompt}

\begin{figure}[H]
  \centering
  \includegraphics[width=\columnwidth]{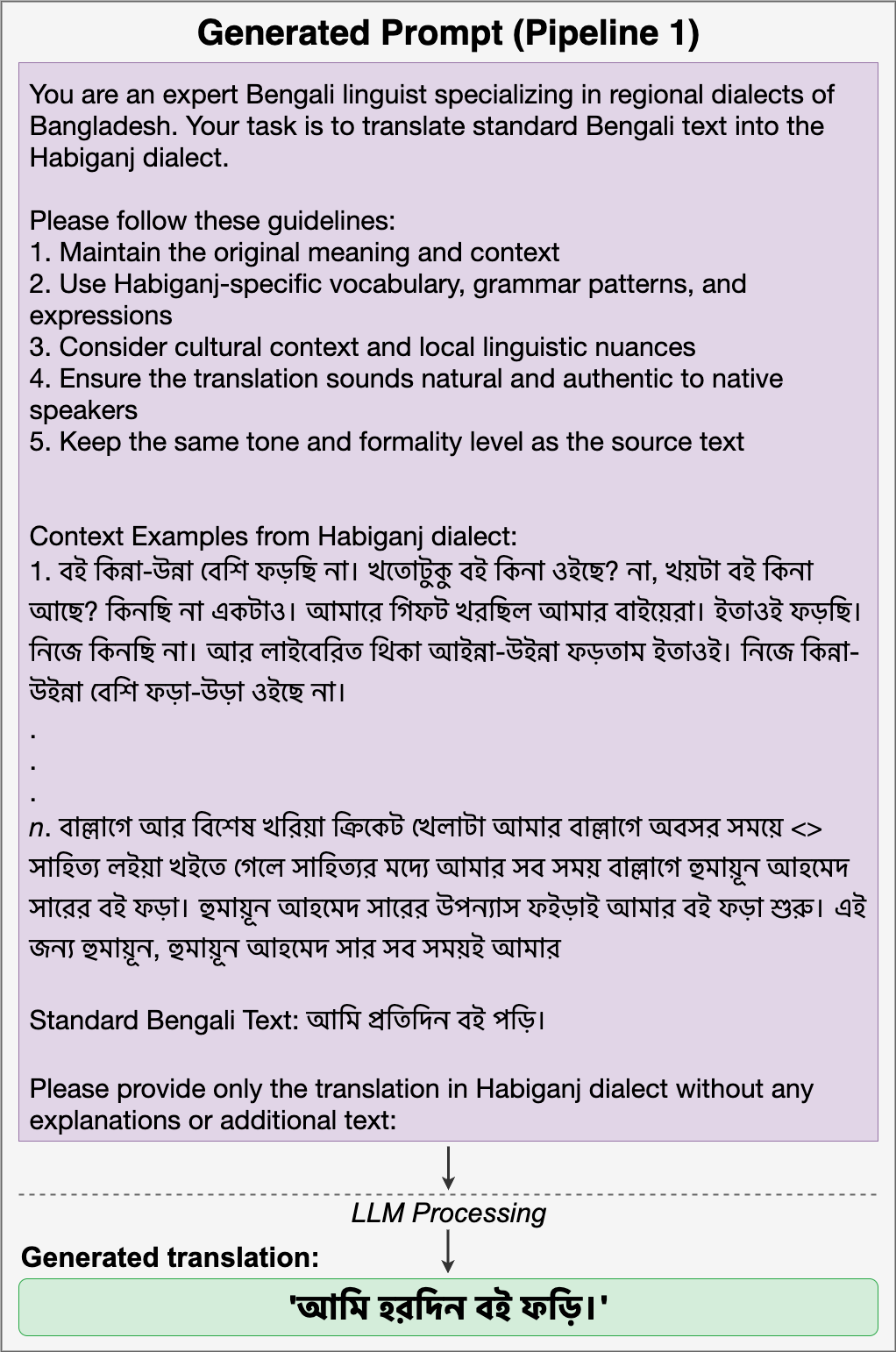}
  
  \caption{A sample prompt that generated while translating a sentence using Pipeline 1 (Transcript-Based Pipeline). The prompt contains up to $n$ retrieved context examples.}
  \label{fig:pipeline1_prompt}
\end{figure}

\subsection{Pipeline 2: Standardized Sentence-Pairs Prompt Example}
\label{sec:pipeline2_prompt}

\begin{figure}[H]
  \centering
  \includegraphics[width=\columnwidth]{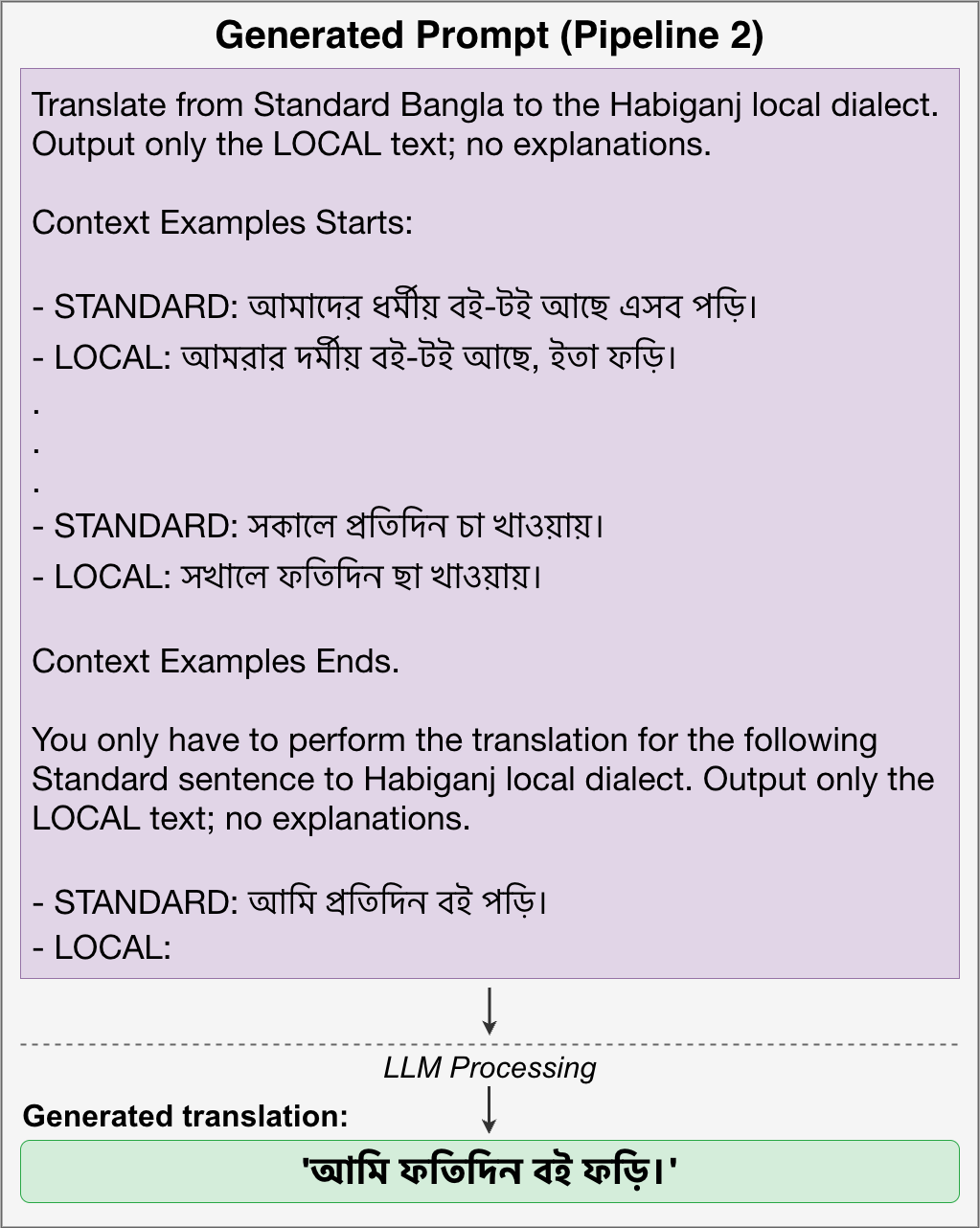}
  \caption{A sample prompt that generated while translating a sentence using Pipeline 2 (Standardized Sentence-Pairs Pipeline). The prompt includes up to $n$ retrieved standard\_bengali:local\_dialect sentence pairs as few-shot examples.}
  \label{fig:pipeline2_prompt}
\end{figure}

\section{Detailed Model Performance Analysis}
\label{sec:detailed_results}

This section provides comprehensive performance breakdowns for all evaluated LLMs across zero-shot, Pipeline 1, and Pipeline 2 conditions. All values are reported as mean $\pm$ standard deviation across all dialects evaluated.

\subsection{Model Comparison Across Conditions}
\label{sec:model_comparison}

\begin{figure}[t]
  \centering
  \includegraphics[width=\columnwidth]{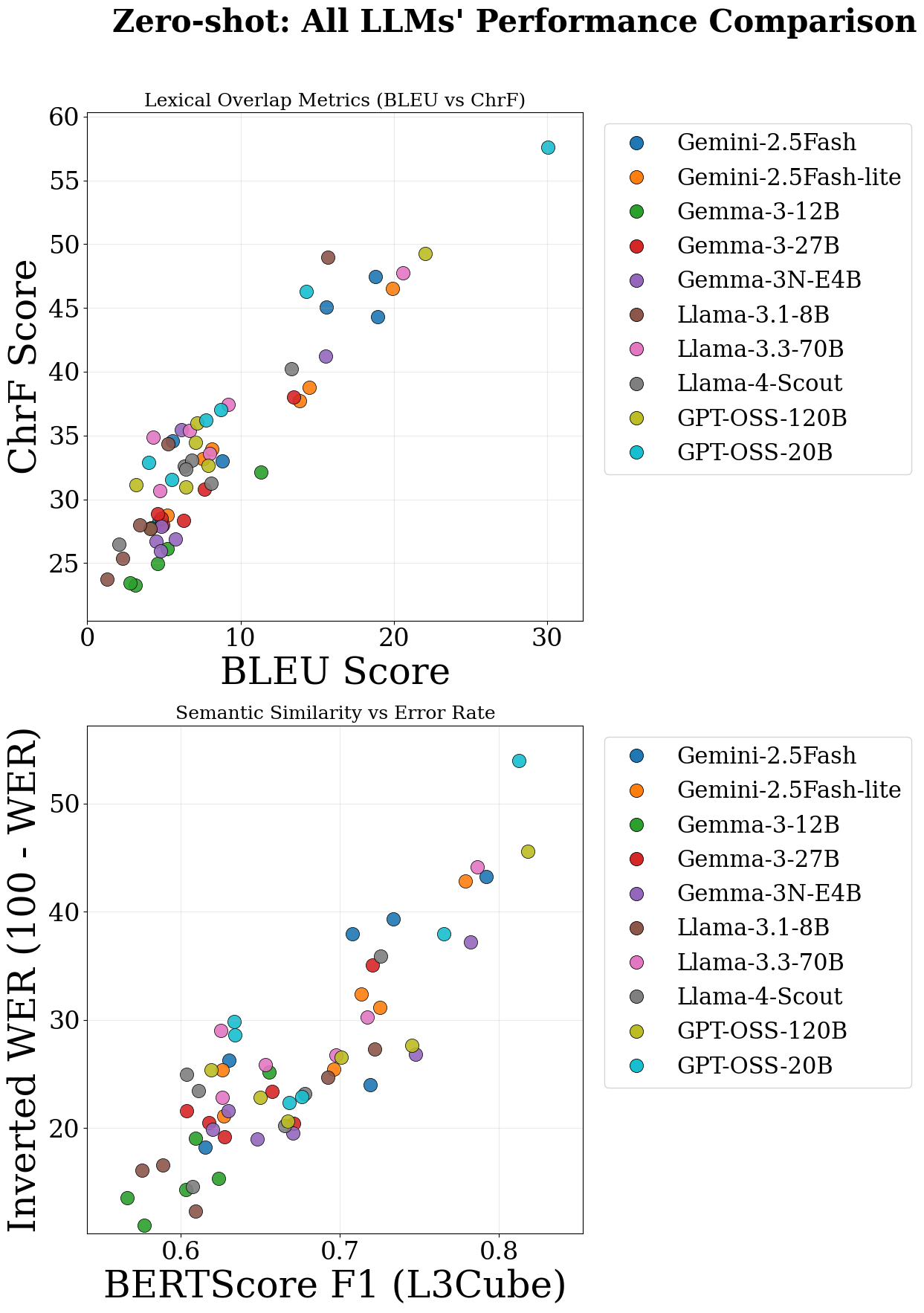}
  \caption{Performance comparison of different LLMs in zero-shot scenario. All models show uniformly poor performance (BLEU 5-12, WER 67-84\%), confirming the necessity of RAG-based approaches for this task.}
  \label{fig:zeroshotall_appendix}
\end{figure}

\begin{figure}[t]
  \centering
  \includegraphics[width=\columnwidth]{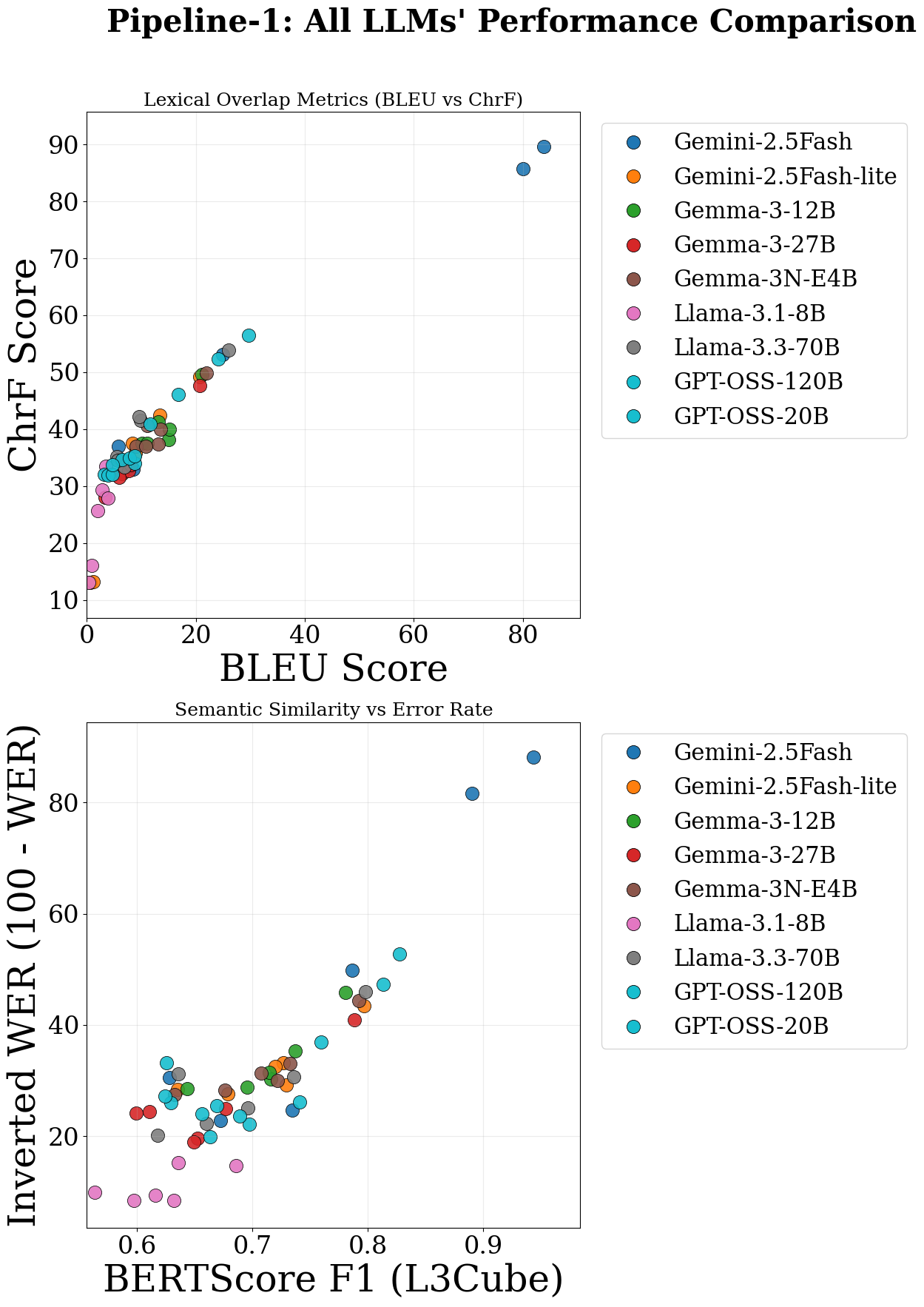}
  \caption{Performance comparison of different LLMs using Pipeline 1 (transcript-based). Gemini-2.5Fash shows notably superior performance (BLEU=34.87, WER=50.39), suggesting stronger ability to infer dialectal patterns from less structured context.}
  \label{fig:pipeline1all_appendix}
\end{figure}

\begin{figure}[t]
  \centering
  \includegraphics[width=\columnwidth]{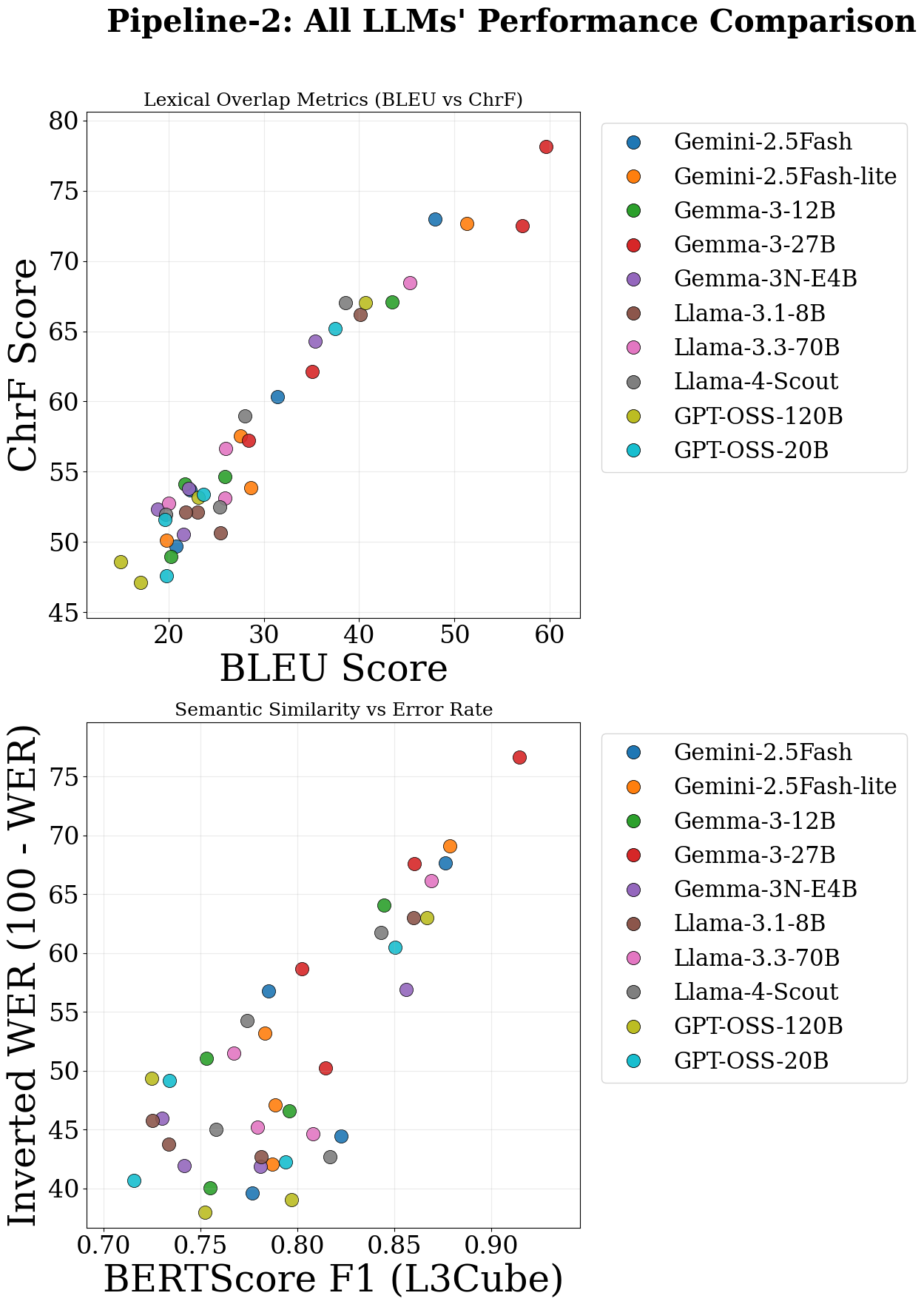}
  \caption{Performance comparison of different LLMs using Pipeline 2 (standardized sentence-pairs). Gemma-3-27B achieves the best overall results (BLEU=45.06, WER=36.70). The superior pipeline narrows the performance gap between models of different sizes.}
  \label{fig:pipeline2all_appendix}
\end{figure}

\subsection{Zero-Shot Performance}
\label{sec:zeroshot_details}

The zero-shot baseline establishes that standard-to-dialect Bengali translation is a challenging task requiring contextual examples. Performance is uniformly low across all models:

\begin{itemize}
  \item \textbf{Best performing models}: Gemini-2.5Fash (BLEU=12.06±6.53, WER=68.47±10.01) and GPT-OSS-20B (BLEU=11.72±9.65, WER=67.37±11.90) show marginally better results but remain far from acceptable translation quality.
  \item \textbf{Smaller models}: Models like Gemma-3-12B (BLEU=5.21±3.14, WER=83.56±5.04) and Llama-3.1-8B (BLEU=5.35±5.26, WER=82.39±7.16) struggle significantly without contextual guidance.
  \item \textbf{Key insight}: Even large 70B+ parameter models fail to produce quality dialectal translations zero-shot, with BLEU scores consistently below 12 and WER above 67\%. This confirms the task's inherent difficulty and data scarcity.
\end{itemize}

\subsection{Pipeline 1 Performance}
\label{sec:pipeline1_details}

Pipeline 1 uses transcript-based context with longer, more descriptive dialectal sentences. Performance improvements over zero-shot are modest for most models, with one notable exception:

\begin{itemize}
  \item \textbf{Gemini-2.5Fash (outlier)}: Achieves BLEU=34.87±37.15 and WER=50.39±29.05, dramatically outperforming all other models in this pipeline. The high standard deviations suggest strong performance on some dialects but inconsistency across others.
  \item \textbf{Other models}: Most models show minimal improvement over zero-shot. For example, Gemma-3-12B improves to BLEU=14.27±3.91 (from 5.21), while GPT-OSS-120B remains at BLEU=9.04±7.69 (barely changed from 8.96).
  \item \textbf{Llama-3.1-8B failure}: This model performs worse than zero-shot (BLEU=2.22±1.39, WER=88.96±3.10), suggesting the transcript-based context format may confuse smaller models lacking sufficient Bengali pretraining.
\end{itemize}

\subsection{Pipeline 2 Performance}
\label{sec:pipeline2_details}

Pipeline 2 provides structured local\_dialect:standard\_bengali sentence pairs, resulting in dramatic and consistent improvements across all models:

\begin{itemize}
  \item \textbf{Top tier models}: Gemma-3-27B leads with BLEU=45.06±15.67 and WER=36.70±11.41, followed closely by Gemini-2.5Fash-lite (BLEU=31.80±13.59, WER=47.14±11.75) and Gemini-2.5Fash (BLEU=30.62±12.51, WER=47.88±12.62).
  \item \textbf{Mid-tier performance}: Models like Llama-3.3-70B (WER=48.12±10.02), Llama-4-Scout (WER=49.07±8.77), and Gemma-3-12B (WER=49.56±10.16) cluster in the 48-50\% WER range, demonstrating solid translation quality.
  \item \textbf{Smaller models competitive}: Llama-3.1-8B (WER=51.18±9.56) and Gemma-3N-E4B (WER=53.33±7.11) become viable options, performing comparably to much larger models like GPT-OSS-120B (WER=52.65±11.64).
  \item \textbf{Performance convergence}: Standard deviations decrease substantially compared to Pipeline 1, indicating more consistent performance across dialects with this approach.
\end{itemize}

\subsection{Key Takeaways}

\begin{enumerate}
  \item \textbf{RAG is essential}: The zero-shot baseline confirms that LLMs lack intrinsic dialectal translation knowledge, regardless of size.
  \item \textbf{Data structure matters more than volume}: Pipeline 2's explicit sentence pairs outperform Pipeline 1's longer transcripts, even with similar data volumes.
  \item \textbf{Smaller models become viable}: With proper context, 8B-12B parameter models can outperform 70B-120B models, democratizing dialectal translation.
  \item \textbf{Model selection depends on pipeline}: Gemini-2.5Fash excels in Pipeline 1, while Gemma-3-27B dominates Pipeline 2, suggesting different architectural strengths.
\end{enumerate}

\section{Comparison with Fine-Tuned Baseline}
\label{sec:khandaker_comparison}

To contextualize our RAG-based approach, we compare our pipelines against the fine-tuned models from \citet{khandaker2025bridging}, who pioneered supervised standard-to-dialect Bengali translation. For this comparison, presented in Table~\ref{tab:khandaker_comparison}, we've taken the best result from our LLM-Pipeline-Dialect combinations, focusing on the Chittagong and Sylhet dialects as these are the only ones where both studies use Word Error Rate (WER) as a common metric. It is important to note the methodological differences: \citet{khandaker2025bridging} utilized supervised fine-tuning of BanglaT5 on parallel corpora, which demands significant training data and resources. In contrast, our approach is fine-tuning-free, relying on retrieval-based in-context learning.

\begin{table}[H]
  \centering
  \small
  \begin{tabular}{lccc}
    \hline
    \textbf{Dialect} & \textbf{Khandaker} & \textbf{Pipeline 1} & \textbf{Pipeline 2} \\
                     & \textbf{et al.} & \textbf{(Best)} & \textbf{(Best)} \\
    \hline
    Chittagong & 70.66 & 71.16 & \textbf{32.37} \\
    Sylhet    & 60.64 & \textbf{18.37} & - \\
    \hline
  \end{tabular}
  \caption{WER (\%) comparison between Khandaker et al.'s fine-tuned BanglaT5 model and our RAG-based pipelines for shared dialects.}
  \label{tab:khandaker_comparison}
\end{table}

\section{Complete Performance Tables}
\label{sec:complete_tables}

This section presents comprehensive performance tables for all three experimental setups: Pipeline 1 (Transcript-Based), Pipeline 2 (Standardized Sentence-Pairs), and Zero-Shot. Each table shows results for all evaluated LLMs across all dialects, with the best performance for each dialect-metric combination highlighted in bold.

\subsection{Pipeline 1: Complete Results}
\label{sec:pipeline1_complete}

\begin{table*}[p]
\centering
\small
\begin{tabular}{llrrrr}
\hline
\textbf{Dialect} & \textbf{Model} & \textbf{BLEU} & \textbf{ChrF} & \textbf{BERTScore F1} & \textbf{WER} $\downarrow$ \\
\hline
Chittagong & gemini-2.5-flash & 8.53 & 32.89 & 0.6722 & 77.18 \\
Chittagong & gemini-2.5-flash-lite & 8.90 & 35.86 & 0.6790 & 72.41 \\
Chittagong & \textbf{gemma-3-12b-it} & \textbf{14.95} & \textbf{38.20} & \textbf{0.6955} & \textbf{71.16} \\
Chittagong & gemma-3-27b-it & 6.47 & 32.20 & 0.6112 & 75.52 \\
Chittagong & gemma-3n-e4b-it & 13.12 & 37.34 & 0.6764 & 71.78 \\
Chittagong & llama-3.1-8b-instant & 2.72 & 29.25 & 0.5974 & 91.49 \\
Chittagong & llama-3.3-70b-versatile & 5.56 & 35.12 & 0.6178 & 79.88 \\
Chittagong & openaigpt-oss-120b & 8.74 & 33.97 & 0.6692 & 74.48 \\
Chittagong & openaigpt-oss-20b & 8.81 & 35.23 & 0.6244 & 72.82 \\
\hline
Comilla & \textbf{gemini-2.5-flash} & \textbf{83.83} & \textbf{89.63} & \textbf{0.9433} & \textbf{11.79} \\
Comilla & gemini-2.5-flash-lite & 0.63 & 13.05 & 0.7270 & 66.81 \\
Comilla & gemma-3-12b-it & 15.15 & 39.96 & 0.7374 & 64.63 \\
Comilla & gemma-3-27b-it & 3.33 & 28.04 & 0.6528 & 80.42 \\
Comilla & gemma-3n-e4b-it & 13.54 & 39.99 & 0.7326 & 66.95 \\
Comilla & llama-3.1-8b-instant & 1.98 & 25.65 & 0.6324 & 91.58 \\
Comilla & llama-3.3-70b-versatile & 6.91 & 33.29 & 0.6960 & 74.95 \\
Comilla & openaigpt-oss-120b & 3.23 & 32.10 & 0.6976 & 77.89 \\
Comilla & openaigpt-oss-20b & 4.71 & 32.01 & 0.6890 & 76.42 \\
\hline
Habiganj & gemini-2.5-flash & 5.82 & 37.03 & 0.6284 & 69.49 \\
Habiganj & gemini-2.5-flash-lite & 8.32 & 37.46 & 0.6354 & 71.61 \\
Habiganj & gemma-3-12b-it & 10.17 & 37.48 & \textbf{0.6435} & 71.40 \\
Habiganj & gemma-3-27b-it & 7.65 & 32.68 & 0.5992 & 75.85 \\
Habiganj & gemma-3n-e4b-it & 8.97 & 36.95 & 0.6329 & 72.46 \\
Habiganj & llama-3.1-8b-instant & 0.93 & 16.11 & 0.5636 & 90.04 \\
Habiganj & llama-3.3-70b-versatile & 9.84 & \textbf{41.55} & 0.6360 & 68.86 \\
Habiganj & openaigpt-oss-120b & 6.46 & 34.63 & 0.6292 & 73.94 \\
Habiganj & \textbf{openaigpt-oss-20b} & \textbf{11.59} & 40.92 & 0.6256 & \textbf{66.74} \\
\hline
Rangpur & gemini-2.5-flash & 6.06 & 34.68 & 0.7344 & 75.32 \\
Rangpur & gemini-2.5-flash-lite & 1.10 & 13.24 & 0.7293 & 70.82 \\
Rangpur & gemma-3-12b-it & 11.10 & 37.46 & 0.7158 & 69.74 \\
Rangpur & gemma-3-27b-it & 8.02 & 33.73 & 0.6770 & 75.11 \\
Rangpur & gemma-3n-e4b-it & 10.76 & 36.93 & 0.7218 & 69.96 \\
Rangpur & llama-3.1-8b-instant & 3.51 & 33.52 & 0.6359 & 84.76 \\
Rangpur & llama-3.3-70b-versatile & 9.59 & 42.24 & 0.7359 & 69.31 \\
Rangpur & openaigpt-oss-120b & 7.87 & 34.96 & 0.7411 & 73.82 \\
Rangpur & \textbf{openaigpt-oss-20b} & \textbf{16.71} & \textbf{46.12} & \textbf{0.7595} & \textbf{63.09} \\
\hline
Sylhet & \textbf{gemini-2.5-flash} & \textbf{80.03} & \textbf{85.78} & \textbf{0.8904} & \textbf{18.37} \\
Sylhet & gemini-2.5-flash-lite & 13.40 & 42.42 & 0.7197 & 67.43 \\
Sylhet & gemma-3-12b-it & 13.12 & 41.24 & 0.7146 & 68.48 \\
Sylhet & gemma-3-27b-it & 5.97 & 31.53 & 0.6491 & 81.00 \\
Sylhet & gemma-3n-e4b-it & 11.11 & 40.60 & 0.7075 & 68.68 \\
Sylhet & llama-3.1-8b-instant & 0.35 & 13.01 & 0.6158 & 90.61 \\
Sylhet & llama-3.3-70b-versatile & 5.59 & 34.57 & 0.6605 & 77.66 \\
Sylhet & openaigpt-oss-120b & 3.81 & 31.89 & 0.6635 & 80.17 \\
Sylhet & openaigpt-oss-20b & 4.71 & 33.73 & 0.6566 & 75.99 \\
\hline
Tangail & gemini-2.5-flash & 24.96 & 53.06 & 0.7866 & 50.21 \\
Tangail & gemini-2.5-flash-lite & 20.74 & 49.16 & 0.7970 & 56.51 \\
Tangail & gemma-3-12b-it & 21.12 & 49.53 & 0.7807 & 54.20 \\
Tangail & gemma-3-27b-it & 20.69 & 47.70 & 0.7883 & 59.03 \\
Tangail & gemma-3n-e4b-it & 21.90 & 49.80 & 0.7924 & 55.67 \\
Tangail & llama-3.1-8b-instant & 3.82 & 27.94 & 0.6861 & 85.29 \\
Tangail & llama-3.3-70b-versatile & 25.98 & 53.90 & 0.7980 & 53.99 \\
Tangail & openaigpt-oss-120b & 24.10 & 52.38 & 0.8135 & 52.73 \\
Tangail & \textbf{openaigpt-oss-20b} & \textbf{29.63} & \textbf{56.54} & \textbf{0.8276} & \textbf{47.27} \\
\hline
\end{tabular}
\caption{Complete Pipeline 1 (Transcript-Based) results for all LLMs across all dialects. Best performance for each dialect-metric combination is shown in bold. LLM names are bolded when they achieve the most wins across all metrics for that dialect.}
\label{tab:pipeline1_complete}
\end{table*}

\subsection{Pipeline 2: Complete Results}
\label{sec:pipeline2_complete}

\begin{table*}[p]
\centering
\small
\begin{tabular}{llrrrr}
\hline
\textbf{Dialect} & \textbf{Model} & \textbf{BLEU} & \textbf{ChrF} & \textbf{BERTScore F1} & \textbf{WER} $\downarrow$ \\
\hline
Chittagong & gemini-2.5-flash & 20.75 & 49.70 & 0.7766 & 60.37 \\
Chittagong & gemini-2.5-flash-lite & 28.58 & 53.85 & 0.7886 & 52.90 \\
Chittagong & gemma-3-12b-it & 20.22 & 48.96 & 0.7551 & 59.96 \\
Chittagong & \textbf{gemma-3-27b-it} & \textbf{57.14} & \textbf{72.52} & \textbf{0.8603} & \textbf{32.37} \\
Chittagong & gemma-3n-e4b-it & 21.57 & 50.51 & 0.7415 & 58.09 \\
Chittagong & llama-3.1-8b-instant & 25.41 & 50.66 & 0.7334 & 56.22 \\
Chittagong & llama-3.3-70b-versatile & 25.93 & 53.12 & 0.7795 & 54.77 \\
Chittagong & llama-4-scout-17b-16e-instruct & 25.38 & 52.46 & 0.7581 & 54.98 \\
Chittagong & openaigpt-oss-120b & 17.00 & 47.09 & 0.7523 & 62.03 \\
Chittagong & openaigpt-oss-20b & 19.73 & 47.60 & 0.7158 & 59.34 \\
\hline
Habiganj & gemini-2.5-flash & 31.43 & 60.33 & 0.7850 & 43.22 \\
Habiganj & gemini-2.5-flash-lite & 27.52 & 57.57 & 0.7831 & 46.82 \\
Habiganj & gemma-3-12b-it & 25.90 & 54.66 & 0.7529 & 48.94 \\
Habiganj & \textbf{gemma-3-27b-it} & \textbf{35.07} & \textbf{62.14} & \textbf{0.8023} & \textbf{41.31} \\
Habiganj & gemma-3n-e4b-it & 22.05 & 53.78 & 0.7303 & 54.03 \\
Habiganj & llama-3.1-8b-instant & 23.01 & 52.10 & 0.7250 & 54.24 \\
Habiganj & llama-3.3-70b-versatile & 25.97 & 56.66 & 0.7673 & 48.52 \\
Habiganj & llama-4-scout-17b-16e-instruct & 27.97 & 58.96 & 0.7742 & 45.76 \\
Habiganj & openaigpt-oss-120b & 23.07 & 53.18 & 0.7249 & 50.64 \\
Habiganj & openaigpt-oss-20b & 23.64 & 53.36 & 0.7341 & 50.85 \\
\hline
Rangpur & gemini-2.5-flash & 22.28 & 53.68 & \textbf{0.8224} & 55.58 \\
Rangpur & gemini-2.5-flash-lite & 19.77 & 50.11 & 0.7871 & 57.94 \\
Rangpur & gemma-3-12b-it & 21.73 & 54.11 & 0.7959 & 53.43 \\
Rangpur & \textbf{gemma-3-27b-it} & \textbf{28.38} & \textbf{57.24} & 0.8144 & \textbf{49.79} \\
Rangpur & gemma-3n-e4b-it & 18.85 & 52.32 & 0.7809 & 58.15 \\
Rangpur & llama-3.1-8b-instant & 21.74 & 52.13 & 0.7813 & 57.30 \\
Rangpur & llama-3.3-70b-versatile & 19.98 & 52.75 & 0.8079 & 55.36 \\
Rangpur & llama-4-scout-17b-16e-instruct & 19.67 & 51.96 & 0.8169 & 57.30 \\
Rangpur & openaigpt-oss-120b & 14.97 & 48.61 & 0.7968 & 60.94 \\
Rangpur & openaigpt-oss-20b & 19.59 & 51.59 & 0.7938 & 57.73 \\
\hline
Tangail & gemini-2.5-flash & 48.00 & 72.99 & 0.8762 & 32.35 \\
Tangail & gemini-2.5-flash-lite & 51.32 & 72.65 & 0.8786 & 30.88 \\
Tangail & gemma-3-12b-it & 43.48 & 67.06 & 0.8447 & 35.92 \\
Tangail & \textbf{gemma-3-27b-it} & \textbf{59.65} & \textbf{78.15} & \textbf{0.9143} & \textbf{23.32} \\
Tangail & gemma-3n-e4b-it & 35.41 & 64.28 & 0.8562 & 43.07 \\
Tangail & llama-3.1-8b-instant & 40.16 & 66.19 & 0.8598 & 36.97 \\
Tangail & llama-3.3-70b-versatile & 45.35 & 68.43 & 0.8689 & 33.82 \\
Tangail & llama-4-scout-17b-16e-instruct & 38.57 & 67.04 & 0.8431 & 38.24 \\
Tangail & openaigpt-oss-120b & 40.69 & 67.02 & 0.8669 & 36.97 \\
Tangail & openaigpt-oss-20b & 37.45 & 65.19 & 0.8503 & 39.50 \\
\hline
\end{tabular}
\caption{Complete Pipeline 2 (Standardized Sentence-Pairs) results for all LLMs across all dialects. Best performance for each dialect-metric combination is shown in bold. LLM names are bolded when they achieve the most wins across all metrics for that dialect.}
\label{tab:pipeline2_complete}
\end{table*}

\subsection{Zero-Shot: Complete Results}
\label{sec:zeroshot_complete}

\begin{table*}[p]
\centering
\small
\begin{tabular}{llrrrr}
\hline
\textbf{Dialect} & \textbf{Model} & \textbf{BLEU} & \textbf{ChrF} & \textbf{BERTScore F1} & \textbf{WER} $\downarrow$ \\
\hline
Chittagong & gemini-2.5-flash & 4.68 & 28.29 & 0.6153 & 81.74 \\
Chittagong & gemini-2.5-flash-lite & 5.22 & 28.75 & 0.6271 & 78.84 \\
Chittagong & gemma-3-12b-it & 4.60 & 24.95 & 0.6033 & 85.68 \\
Chittagong & gemma-3-27b-it & 4.91 & 27.99 & 0.6177 & 79.46 \\
Chittagong & gemma-3n-e4b-it & 4.48 & 26.71 & 0.6200 & 80.08 \\
Chittagong & llama-3.1-8b-instant & 4.12 & 27.73 & 0.5889 & 83.40 \\
Chittagong & llama-3.3-70b-versatile & 4.75 & 30.66 & 0.6262 & 77.18 \\
Chittagong & llama-4-scout-17b-16e-instruct & 6.33 & 32.60 & 0.6113 & 76.56 \\
Chittagong & openaigpt-oss-120b & 6.44 & 30.95 & \textbf{0.6500} & 77.18 \\
Chittagong & \textbf{openaigpt-oss-20b} & \textbf{7.72} & \textbf{36.19} & 0.6338 & \textbf{71.37} \\
\hline
Comilla & \textbf{gemini-2.5-flash} & \textbf{18.94} & \textbf{44.34} & \textbf{0.7336} & \textbf{60.63} \\
Comilla & gemini-2.5-flash-lite & 13.85 & 37.74 & 0.7253 & 68.84 \\
Comilla & gemma-3-12b-it & 5.24 & 26.10 & 0.6238 & 84.63 \\
Comilla & gemma-3-27b-it & 4.83 & 28.49 & 0.6711 & 79.58 \\
Comilla & gemma-3n-e4b-it & 5.76 & 26.86 & 0.6704 & 80.42 \\
Comilla & llama-3.1-8b-instant & 3.44 & 27.99 & 0.6383 & 91.37 \\
Comilla & llama-3.3-70b-versatile & 7.97 & 33.59 & 0.6976 & 73.26 \\
Comilla & llama-4-scout-17b-16e-instruct & 8.09 & 31.24 & 0.6652 & 79.79 \\
Comilla & openaigpt-oss-120b & 7.04 & 34.49 & 0.7010 & 73.47 \\
Comilla & openaigpt-oss-20b & 5.53 & 31.55 & 0.6760 & 77.05 \\
\hline
Habiganj & gemini-2.5-flash & 5.54 & 34.58 & 0.6304 & 73.73 \\
Habiganj & gemini-2.5-flash-lite & 7.55 & 33.19 & 0.6263 & 74.58 \\
Habiganj & gemma-3-12b-it & 3.15 & 23.25 & 0.5664 & 86.44 \\
Habiganj & gemma-3-27b-it & 6.30 & 28.34 & 0.6035 & 78.39 \\
Habiganj & gemma-3n-e4b-it & 4.81 & 27.86 & 0.6300 & 78.39 \\
Habiganj & llama-3.1-8b-instant & 2.32 & 25.35 & 0.5755 & 83.90 \\
Habiganj & llama-3.3-70b-versatile & 6.67 & 35.41 & 0.6250 & 70.97 \\
Habiganj & llama-4-scout-17b-16e-instruct & 6.84 & 33.04 & 0.6034 & 75.00 \\
Habiganj & openaigpt-oss-120b & 7.89 & 32.64 & 0.6191 & 74.58 \\
Habiganj & \textbf{openaigpt-oss-20b} & \textbf{8.72} & \textbf{37.04} & \textbf{0.6334} & \textbf{70.13} \\
\hline
Rangpur & gemini-2.5-flash & 8.81 & 33.03 & 0.7193 & 75.97 \\
Rangpur & gemini-2.5-flash-lite & \textbf{14.48} & 38.77 & 0.7135 & 67.60 \\
Rangpur & gemma-3-12b-it & 4.14 & 27.78 & 0.6092 & 80.90 \\
Rangpur & gemma-3-27b-it & 7.65 & 30.79 & 0.6575 & 76.61 \\
Rangpur & gemma-3n-e4b-it & 6.15 & 35.47 & 0.7476 & 73.18 \\
Rangpur & llama-3.1-8b-instant & 5.25 & 34.32 & 0.6923 & 75.32 \\
Rangpur & llama-3.3-70b-versatile & 9.22 & 37.45 & 0.7173 & 69.74 \\
Rangpur & llama-4-scout-17b-16e-instruct & 6.44 & 32.37 & 0.6780 & 76.82 \\
Rangpur & openaigpt-oss-120b & 7.15 & 35.96 & 0.7453 & 72.32 \\
Rangpur & \textbf{openaigpt-oss-20b} & 14.30 & \textbf{46.29} & \textbf{0.7656} & \textbf{62.02} \\
\hline
Sylhet & \textbf{gemini-2.5-flash} & \textbf{15.61} & \textbf{45.06} & \textbf{0.7077} & \textbf{62.00} \\
Sylhet & gemini-2.5-flash-lite & 8.11 & 33.91 & 0.6960 & 74.53 \\
Sylhet & gemma-3-12b-it & 2.78 & 23.42 & 0.5770 & 88.94 \\
Sylhet & gemma-3-27b-it & 4.58 & 28.89 & 0.6274 & 80.79 \\
Sylhet & gemma-3n-e4b-it & 4.80 & 25.93 & 0.6480 & 81.00 \\
Sylhet & llama-3.1-8b-instant & 1.27 & 23.71 & 0.6092 & 87.68 \\
Sylhet & llama-3.3-70b-versatile & 4.29 & 34.86 & 0.6534 & 74.11 \\
Sylhet & llama-4-scout-17b-16e-instruct & 2.07 & 26.50 & 0.6073 & 85.39 \\
Sylhet & openaigpt-oss-120b & 3.20 & 31.12 & 0.6674 & 79.33 \\
Sylhet & openaigpt-oss-20b & 4.02 & 32.91 & 0.6682 & 77.66 \\
\hline
Tangail & gemini-2.5-flash & 18.80 & 47.47 & 0.7918 & 56.72 \\
Tangail & gemini-2.5-flash-lite & 19.90 & 46.50 & 0.7790 & 57.14 \\
Tangail & gemma-3-12b-it & 11.34 & 32.14 & 0.6553 & 74.79 \\
Tangail & gemma-3-27b-it & 13.47 & 38.01 & 0.7205 & 64.92 \\
Tangail & gemma-3n-e4b-it & 15.57 & 41.20 & 0.7822 & 62.82 \\
Tangail & llama-3.1-8b-instant & 15.70 & 48.95 & 0.7217 & 72.69 \\
Tangail & llama-3.3-70b-versatile & 20.59 & 47.78 & 0.7864 & 55.88 \\
Tangail & llama-4-scout-17b-16e-instruct & 13.31 & 40.21 & 0.7256 & 64.08 \\
Tangail & openaigpt-oss-120b & 22.05 & 49.29 & \textbf{0.8183} & 54.41 \\
Tangail & \textbf{openaigpt-oss-20b} & \textbf{30.05} & \textbf{57.60} & 0.8125 & \textbf{46.01} \\
\hline
\end{tabular}
\caption{Complete Zero-Shot results for all LLMs across all dialects. Best performance for each dialect is shown in bold.}
\label{tab:zeroshot_complete}
\end{table*}

\end{document}